\documentclass{article}
\usepackage{spconf,amsmath,graphicx}
\usepackage{multirow}
\usepackage{tabularx}
\usepackage[normalem]{ulem}
\useunder{\uline}{\ul}{}
\usepackage{booktabs}
\usepackage{colortbl}
\usepackage{adjustbox}
\usepackage[misc]{ifsym}

\newcommand{\ie}{\textit{i}.\textit{e}. }
\title{CSCNet: Class-specified Cascaded Network for Compositional Zero-shot Learning}
\name{Yanyi Zhang~$^{1}$, Qi Jia~$^{1}$, Xin Fan~$^{1}$, Yu Liu~$^{1*}$, Ran He~$^{2*}\thanks{*equal corresponding author.}$
}
\address{$^{1}$~International School of Information Science and Engineering, Dalian University of Technology, China \\
$^{2}$~MAIS\&CRIPAC, Institute of Automation, Chinese Academy of Sciences, China
}
\begin{document}
%\ninept
%
\maketitle
\begin{abstract}
% Compositional Zero-shot Learning (CZSL) aims to enable machines to recognize novel attribute-object compositions based on foregone knowledge.
% Existing methods address this task by disentangling visual features of attributes and objects, however, they lose sight of the importance of attribute-object dependency for disentangled representation learning.
% To this end, we propose a novel attribute-object disentangled framework namely Class-specified Cascaded Network (CSCNet).
% which constructs Attribute-to-Object and Object-to-Attribute disentangled branches.
% Each branch performs the prediction of one primitive independently and then specifies the prediction as prior knowledge for identifying another primitive. 
% Notably, we devise a learnable score classifier to estimate the prediction probabilities instead of the commonly-used cosine score. 
% In addition, a composition branch is also integrated to model the primitives as a whole. 
% Extensive experiments on two real-world datasets show our method is superior to previous competitive methods.
Attribute and object (A-O) disentanglement is a fundamental and critical problem for Compositional Zero-shot Learning (CZSL), whose aim is to recognize novel A-O compositions based on foregone knowledge. Existing methods based on disentangled representation learning lose sight of the contextual dependency between the A-O primitive pairs. Inspired by this, we propose a novel A-O disentangled framework for CZSL, namely Class-specified Cascaded Network (CSCNet). The key insight is to firstly classify one primitive
and then specifies the predicted class as a priori for guiding another primitive recognition in a cascaded fashion. To this end, CSCNet constructs Attribute-to-Object and Object-to-Attribute cascaded branches,
in addition to a composition branch modeling the two primitives as a whole. Notably, we devise a parametric classifier (ParamCls) to improve the matching between visual and semantic embeddings. 
By improving the A-O disentanglement, our framework achieves
superior results than previous competitive methods.

\end{abstract}

\section{Introduction}
\label{sec:intro}
\begin{figure}[t]
  \centering
  \includegraphics[width=\columnwidth]{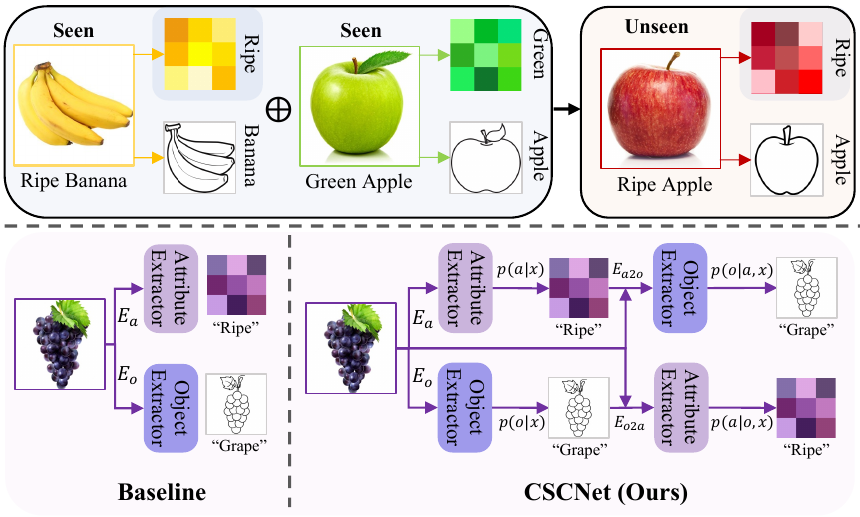}
  \vspace{-0.5cm}
  \caption{Top: concept illustration of CZSL, where a novel composition ``Ripe Apple'' recomposes attribute and object primitives learned from known compositions ``Ripe Banana'' and ``Green Apple''.
  Bottom: comparing the A-O disentanglement between the baseline method and CSCNet we propose. Note that, we omit the composition branch for brevity.
   }
   \label{fig: first}
\end{figure}

The hallmark of human cognitive system is compositionality, which enables us to effortlessly reason about unknown categories by recomposing existing concepts.
% For instance, let's consider a scenario where an individual encounters a ``Square Watermelon'' for the first time. Despite never having seen a square watermelon before, the person can readily identify it by drawing upon their knowledge of a ``Square Box'' and an ``Oval Watermelon''.
% With the rapid progress of deep learning, equipping machine vision systems with the ability to compositional inference has become an essential challenge. To this end, 
Inspired by this, Compositional Zero-shot Learning~(CZSL)~\cite{RedWine, AFGC, causal} is proposed to make the machines recognize novel attribute-object (A-O) compositions by acquiring knowledge from known ones, as shown in the top of Fig.~\ref{fig: first}.

The core in CZSL is how to isolate attribute and object information from a unified visual feature.
Most methods~\cite{Dual-Stream, punan4, IVR, OADis, DECA, BMP-Net, punan3}, 
address this problem through disentangled representation learning, which builds two independent primitive recognition branches separately, see ``Baseline'' in Fig.~\ref{fig: first}.
% adopt the disentangled approach to enhance the model's performance by incorporating independent primitive recognition branches as illustrated in the lower left of Fig.~\ref{fig: first} which is also our baseline. 
However, these methods overlook the contextual dependency between A-O during primitive modeling, leading to less generalizability on novel and unseen compositions. 
% Recently, SeCoNet~\cite{SeCoNet} proposed to utilize the decoupled visual feature maps of the first primitive to guide the second primitive decoupling network, facilitating the inducement between them. However, their inducement is only on visual embeddings and overly restrictive, resulting in less accurate prediction of the second primitive.
% Another work similar to ours, called CANet~\cite{CANet}, proposed to use the semantic embeddings of the object classification results to induce the semantic embeddings of attributes, whereas it neglects the interplay between visual and semantic modalities and object-to-attribute guidance, potentially limiting its overall effectiveness.
Recently, SeCoNet~\cite{SeCoNet} proposed to utilize the decoupled feature maps of the first primitive to guide the second primitive decoupling network, facilitating the inducement between them. However, their inducement is overly restrictive, resulting in less accurate prediction of the second primitive.
Another work similar to ours, called CANet~\cite{CANet}, proposed to leverage object classification results to induce the semantic embeddings of attributes, whereas it neglects the contextuality of the object. For instance, the visual features of potato in mashed potato differ from those of peeled potato.

\begin{figure*}[t]
  \centering
  \includegraphics[width=\textwidth]{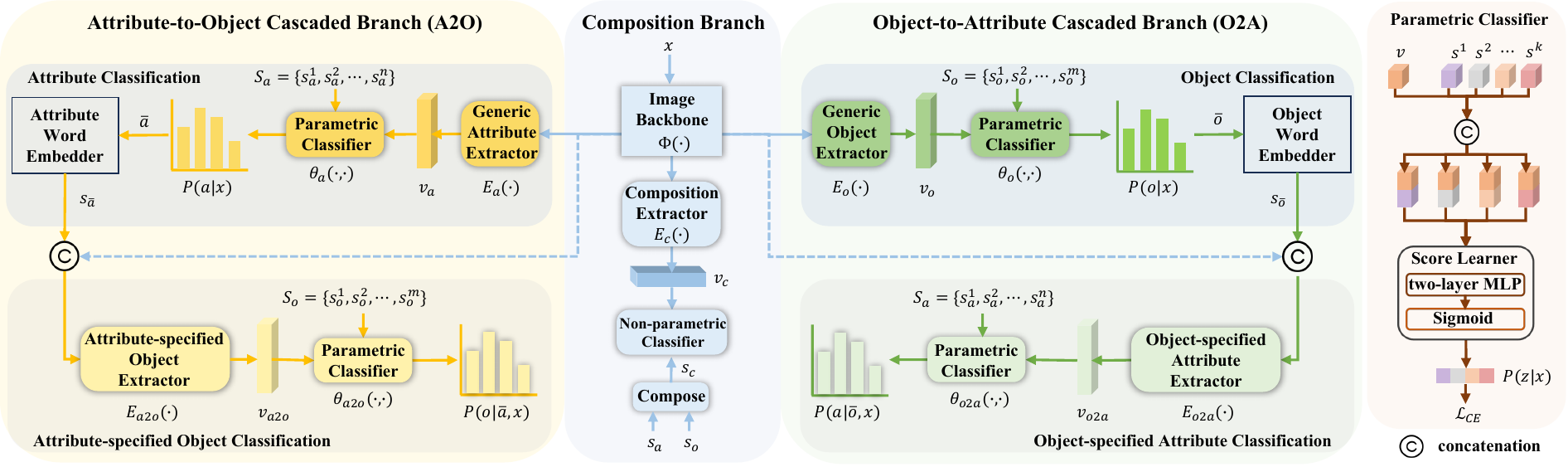}
  \vspace{-0.8cm}
   \caption{Architecture of our Class-specified Cascaded Network (CSCNet) for CZSL.}
   \label{fig:framework}
\end{figure*}

To conquer the limitations above, we propose 
a novel A-O disentangled framework termed 
Class-specified Cascaded Network (CSCNet), as sketched in Fig.~\ref{fig: first}.
Different from previous methods, we exploit class-specified guidance
to model both attribute-to-object and object-to-attribute dependency simultaneously.
Concretely, the model constructs an attribute-to-object (A2O) cascaded branch, 
where it firstly classifies the attribute and then specifies the predicted class as a priori to guide the object recognition.
% where it firstly classifies the attribute and then specifies the predicted class as a priori to guide the object recognition, \ie the semantic embeddings of attributes guide the visual embeddings of objects.
A contrary pipeline takes place in a object-to-attribute (O2A) cascaded branch.
Notably, we devise a parametric classifier (ParamCls) for matching visual-semantic embeddings in A2O and O2A branches.
Advantageously, our ParamCls develops a score learner to learn optimal
matching scores, instead of using the traditional non-parametric manner based cosine score. 
In addition, a composition branch is also available for modeling the two primitives as a whole. 
In summary, our contributions are three-fold: 
1) we develop a novel A-O disentanglement framework modeling contextual dependency with class-specified guidance;
2) we design a parametric classifier to learn optimal matching scores between visual and semantic embeddings;
3) extensive results on two datasets demonstrate CSCNet achieves superior performance compared to previous competitive methods.

\section{Methodology}
\label{sec:method}
\textbf{Overview.}
Fig.~\ref{fig:framework} depicts the architecture of CSCNet for CZSL. Given an image instance, we extract the image feature with a pre-trained backbone. The feature is then fed into the attribute-to-object cascaded branch (A2O), object-to-attribute cascaded branch (O2A), and composition branch. A2O firstly classifies the attribute and then specifies the predicted class as a priori to guide the object classification. Likewise, O2A classifies objects and then uses the results to guide attribute classification. The composition branch classifies A-O pairs as a whole, so as to capture intricate A-O entanglement. 
The three branches can be trained jointly.
% During the classification process, we use ParamCls to generate the classification score actively, enabling dense cross-modal interactions between visual features and semantic features.

\subsection{Attribute-to-Object Cascaded Branch (A2O)}
% Sequential Classification Branches consist of two branches: the Attribute-Prioritized Sequential Classification Branch ($a$→$o$, APSC) and the Object-Prioritized Sequential Classification Branch ($o$→$a$, OPSC).
% Each branch is a two-step prediction, \ie Straightforward Predictions and Contingent Predictions. 
% For APSC, it initially focuses on recognizing attributes in isolation, referred to as Straightforward Predictions.
% Subsequently, in the context of recognizing attributes, APSC proceeds to recognize objects, which is known as Contingent Predictions.
% The same goes for OPSC.

\textbf{Attribute Classification.}
After obtaining the image feature $\Phi(x)$ extracted by an image backbone $\Phi(\cdot)$, we feed it into a generic attribute extractor $E_a(\cdot)$. As a result, a disentangled attribute visual feature $v_a$ is achieved by
$v_a=E_a(\Phi(x))$.
% \begin{equation}
% \begin{aligned}
%     v_a=E_a(\Phi(x)).
% \end{aligned}
% \end{equation}
On the other hand, we have the ground-truth attribute labels
and capture their semantic embeddings from a language encoder like word2vec~\cite{w2v}, denoted as $S_a = \left \{  s^1_a,s^2_a,\dots s^n_a \right \}$, where $S_a$ represents the semantic features of all attributes and $n$ is the number of attribute classes.
For attribute classification, it is important to compute the matching scores between visual and semantic embeddings.
In general, existing methods employ a simple non-parametric manner based on cosine scores. 
On the contrary, we propose to devise a \textbf{parametric classifier (ParamCls)}
which leverages a score learner to make the matching scores be learnable, together with other network parameters. 
As shown in the right of Fig.~\ref{fig:framework}, we concatenate $v_a$ with each $s^i_a$ and then feed them into a score learner, which consists of a two-layer Multilayer Perceptron (MLP) followed by a Sigmoid function.
Formally, the computation in ParamCls is given by 
\begin{equation}
\begin{aligned}
    \theta_a(v_a, S_a) = {\rm Sigmoid}({\rm MLP}({\rm Concat}(v_a, S_a))),
\end{aligned}
\end{equation}
where $\theta_a(\cdot,\cdot)$ outputs the attribute classification probabilities with ParamCls. 
Based on the matching scores, we compute the cross-entropy loss $\mathcal{L}_{a}$ for attribute classification:
\begin{equation}
\begin{aligned}
    % \mathcal{L}_{a} = - \log P(a \mid x)= -\log \theta_a(v_a, S_a),
    \mathcal{L}_{a} = - y_{a} \log P(a \mid x) = - y_{a} \log \theta_a(v_a, S_a),
\end{aligned}
\end{equation}
where one-hot vector $y_{a}$ represents the attribute ground-truth of $x$.
Note that, we omit the index $i$ in $\{(x^{i},y_a^{i})\}_{i=1}^{N}$ for brevity.
After estimating the predicted attribute label $\bar{a}$,
we further capture its corresponding semantic embedding $s_{\bar{a}}$ via
an attribute word embedder, for example $s_{\bar{a}} ={\rm word2vec}(\bar{a})$. Importantly, $s_{\bar{a}}$ will guide
the following object classification.

\textbf{Attribute-specified Object Classification.} 
To grasp the contextual dependency between attributes and objects, we specify the predicted attribute label as a priori to guide the object classification. Intuitively, when the attribute label is provided,
it becomes easier to predict the remaining object label.
To make it, we firstly concatenate the semantic embedding $s_{\bar{a}}$
with the image feature $\Phi(x)$. Then we pass the resulting feature
to a new attribute-specified object extractor $E_{a2o}(\cdot)$
and derive object visual feature $v_{a2o}$:
\begin{equation}
\begin{aligned}
    v_{a2o} = E_{a2o}({\rm Concat}(\Phi(x), s_{\bar{a}})).
\end{aligned}
\end{equation}

% where $v_{a2o}$ is considered the visual features of object guided by prior knowledg as it merge the classification results from previous attribute classification. 
Likewise, we proceed to perform object classification by learning a new ParamCls $\theta_{a2o}(\cdot,\cdot)$ and calculate the cross-entropy loss $\mathcal{L}_{a2o}$:
\begin{equation}
\begin{aligned}
    \mathcal{L}_{a2o} = -y_{o}\log P(o \mid \bar{a}, x)= -y_{o}\log \theta_{a2o}(v_{a2o},S_o),
\end{aligned}
\end{equation}
where $P(o \mid \bar{a}, x)$ is the classification probabilities of an object given attribute prediction $\bar{a}$,
one-hot vector $y_{o}$ indicates the object ground-truth,
and $S_o = \left \{  s^1_o,s^2_o,\dots s^m_o \right \}$ denotes the semantic embeddings of all objects.
In a nutshell, the tasks of attribute and object classification
are linked seamlessly through a cascaded fashion.

\subsection{Object-to-Attribute Cascaded Branch (O2A)}
Similar to A2O, we first perform object classification and then proceed to attribute classification.
Specifically, we decouple the object visual features $v_o$ by a generic object extractor $E_o(\cdot)$, \ie $v_o=E_o(\Phi(x))$.
The loss for this object classification task is formulated as:
\begin{equation}
\begin{aligned}
    \mathcal{L}_{o} = -y_{o}\log P(o \mid x)= -y_{o}\log \theta_o(v_o, S_o).
\end{aligned}
\end{equation}
%where $\theta_o(\cdot, \cdot)$ is the operation in ParamCls.

Subsequently, we convert the object classification result $\bar o$ to semantic embedding $s_{\bar o}$ and specify it as a priori for the following attribute classification. We merge $s_{\bar o}$ with $\Phi(x)$ for deriving attribute embedding $v_{o2a}$ from object-specified attribute extractor $E_{o2a}(\cdot)$. Finally, we match $v_{o2a}$ with $S_a$ by using ParamCls $\theta_{o2a}(\cdot, \cdot)$ and achieve the loss cost $\mathcal{L}_{o2a}$:
\begin{equation}
\begin{aligned}
    \mathcal{L}_{o2a} = -y_{a}\log P(a \mid \bar{o}, x)= -y_{a}\log \theta_{o2a}(v_{o2a}, S_a).
\end{aligned}
\end{equation}

\subsection{Composition Branch}
Although we have performed precise classification of individual primitives, it is still necessary to learn the composition as a whole. This is because there is an intricate entanglement between A-O, and modeling this entanglement is crucial for preserving the contextuality.
Hence, we encode the image feature $\Phi(x)$ to obtain a composition visual features $v_c = E_c(\Phi(x))$,
% \begin{equation}
% \begin{aligned}
%     v_c = E_c(\Phi(x)),
% \end{aligned}
% \end{equation}
where $E_c(\cdot)$ denotes a composition extractor. Then, we compose the primitive semantic embeddings (\ie $s_a$ and $s_o$) into a composition semantic embedding $s_c$ by
\begin{equation}
\begin{aligned}
    s_c = {\rm Compose}(s_a, s_o) = {\rm MLP}({\rm Concat}(s_a, s_o)).
\end{aligned}
\end{equation}

Subsequently, we utilize a non-parametric classifier (NonPaCls) based on cosine score, for matching $v_{c}$ and $s_{c}$.
We need to note that ParamCls is not suitable for composition classification. The reason is that the test set contains some unseen compositions that cannot be learned by ParamCls in advance during training. 
Using ParamCls for composition classification may lead the model to 
being biased on seen compositions during inference.
Differently, since all the objects and attributes have been seen at training phase for CZSL, ParamCls is undoubtedly suitable for attribute and object classification.
Finally, the composition classification loss $\mathcal{L}_{c}$ is
\begin{equation}
\begin{aligned}
    \mathcal{L}_c = -y_{c}\log P(c \mid x) = -\log \frac{\exp \left\langle v_{c}, s_{c}\right\rangle}{\sum_{s_{\hat{c}} \in S_c} \exp \left\langle v_{c}, s_{\hat{c}}\right\rangle},
\end{aligned}
\end{equation}
where $\left\langle \cdot, \cdot \right\rangle$ denotes cosine similarity and $S_c$ includes the semantic embeddings of all compositions.

\subsection{Training and Inference}
%\textbf{Objectives.}
All the three branches in CSCNet can be trained jointly.
The total loss is a combination of five loss terms as follows:
\begin{equation}
\begin{aligned}
    \mathcal{L}_{total} = \alpha (\mathcal{L}_{a} + \mathcal{L}_{o} + \mathcal{L}_{a2o} + \mathcal{L}_{o2a}) + \mathcal{L}_{c},
\end{aligned}
\end{equation}
where $\alpha$ weighs the loss cost in A2O and O2A. 
During inference, we integrate the probabilities from the three branches. The final score is calculated by the formula below:
\begin{equation}
\begin{aligned}
    {\rm Score} = \beta(P(a\mid x)P(o\mid \bar{a}, x) + P(o\mid x)P(a\mid \bar{o}, x)) \\
    + (1-\beta)P(c\mid x),
\end{aligned}
\label{equ: score}
\end{equation}
where $\beta$ is a trade-off hyper-parameter.

\begin{table}[t]
\centering
\caption{Comparison with state-of-the-art results on MIT-States and C-GQA. The best results are shown \textbf{in bold}.}
\begin{adjustbox}{width=\columnwidth}
\begin{tabular}{lccccccccc}
\toprule
Dataset~→ & \multicolumn{4}{c}{MIT-States} & \multicolumn{4}{c}{C-GQA} \\
\cmidrule(lr){2-5} \cmidrule(lr){6-9}
Method~↓ & AUC & HM & Seen & Unseen & AUC & HM & Seen & Unseen \\
\midrule
AttrAsOp~\cite{AttrAsOp} & 1.6 & 9.9 & 14.3 & 17.4 & 0.7 & 5.9 & 17.0 & 5.6 \\
TMN~\cite{TaskDriven} & 2.9 & 13.0 & 20.2 & 20.1 & 1.1 & 7.5 & 23.1 & 6.5 \\
SymNet~\cite{Symnet} & 3.0 & 16.1 & 24.4 & 25.2 & 2.1 & 11.0 & 26.8 & 10.3 \\
SeCoNet~\cite{SeCoNet} & 4.0 & 17.4 & 26.0 & 24.8 & 1.3 & 7.7 & 21.8 & 7.8 \\
CompCos~\cite{Compcos} & 4.5 & 16.4 & 25.3 & 24.6 & 2.6 & 12.4 & 28.1 & 11.2 \\
CGE~\cite{CGE} & 5.1 & 17.2 & 28.7 & 25.3 & 2.3 & 11.4 & 28.1 & 10.1 \\
IVR~\cite{IVR} & - & - & - & - & 2.2 & 10.9 & 27.3 & 10.0 \\
SRA~\cite{SRA} & 5.2 & 17.9 & 29.2 & 24.0 & 2.2 & 11.0 & 26.4 & 10.8 \\
SCEN~\cite{SCEN} & 5.3 & \textbf{18.4} & 29.9 & 25.2 & 2.9 & 12.4 & 28.9 & 12.1 \\
DeCa~\cite{DECA} & 5.3 & 18.2 & 29.8 & 25.2 & - & - & - & - \\
CANet~\cite{CANet} & 5.4 & 17.9 & 29.0 & \textbf{26.2} & 3.3 & \textbf{14.5} & 30.0 & 13.2 \\
\midrule
%Our baseline & 5.1 & 17.7 & 28.5 & 25.3 & 3.2 & 13.7 & 29.9 & 13.3 \\
CSCNet (Ours) & \textbf{5.7} & \textbf{18.4} & \textbf{30.0} & \textbf{26.2} & \textbf{3.4} & 14.4 & \textbf{30.4} & \textbf{13.4} \\
%Improvements & \textbf{+0.6} & \textbf{+0.7} & \textbf{+1.5} & \textbf{+0.9} & \textbf{+0.2} & \textbf{+0.7} & \textbf{+0.5} & \textbf{+0.1} \\
\bottomrule
\end{tabular}
\end{adjustbox}
\label{tab: compare}
\end{table}

\section{Experiments}
\subsection{Dataset and Details}
We conduct experiments on two datasets, MIT-States~\cite{MIT} and C-GQA~\cite{CGE}. 
% MIT-States comprises 245 objects and 115 attributes, which form a total of 1262 seen compositions and 700 unseen compositions. C-GQA is regarded as one of the most challenging datasets in CZSL, which consists of 413 attributes and 674 objects, making up 5592 seen compositions and 1963 unseen compositions.
CZSL is commonly evaluated by four metrics: AUC, HM, Seen, and Unseen.
Seen and Unseen metrics quantify the model's maximum predictive capability over seen and unseen compositions, while AUC and HM metrics assess the holistic performance~\cite{ADE}.
Following the literature, we employ ResNet-18~\cite{resnet} pre-trained on ImageNet~\cite{ImageNet} as the image backbone. Each of the feature extractors is a two-layer MLP, without sharing the parameters. 
The semantic features are parameterized by word2vec+fasttext~\cite{w2v,fasttext} for MIT-States and by word2vec for C-GQA.
CSCNet is trained using the Adam optimizer~\cite{adam2015} with a learning rate of $5.0 \times 10^{-5}$ for both datasets with 300 epochs on MIT-States and 200 epochs on C-GQA. The batch size is 256 on MIT-States and 128 on C-GQA. $\alpha$ is set to 4 on both datasets. $\beta$ is 0.1 on MIT-States and 0.2 on C-GQA.
Experiments were conducted on a single NVIDIA RTX 3090 GPU card.

\subsection{Comparison with State-of-the-arts}
We provide a comprehensive comparison with competitive methods including 
SecoNet~\cite{SeCoNet} and CANet~\cite{CANet}.
Table~\ref{tab: compare} reports the compared results on two datasets. 
Overall, CSCNet achieves the best results across all metrics for both datasets, except 0.1 lower than CANet on HM for C-GQA. 
Notably, our AUC results, that is 5.7 in MIT-States and 3.4 in C-GQA, represent 5.6\% and 3.0\% gains over previous state-of-the-art methods. Besides, CSCNet exhibits significant improvements on Seen and Unseen metrics, surpassing existing methods with a consistent margin. 
% Although CSCNet achieves 14.4 on HM in C-GQA, the result is comparable to state-of-the-art performance.
% Notably, the proposed method exhibits a higher performance improvement on the C-GQA dataset, indicating its robustness for larger-scale datasets.
% The performance improvement can be credited to two elements of our proposed method: the ``Sequential Classification Branches'' and the ``Cross Encoder Classifier''. The former enables the model to effectively model primitives within the context of Contextuality and Compositionality, resulting in more accurate representations. The deep interaction between cross-modal features alleviates the modality gap on account of the latter.  The integration of two components leads to the observed performance improvements.

\begin{table}[t]
\centering
% \vspace{-0.1cm}
\caption{Ablation study on A2O and O2A in C-GQA.}
\begin{adjustbox}{width=0.85\columnwidth}
\begin{tabular}{lcccc}
\toprule
%\multicolumn{1}{l}{Dataset~→} & \multicolumn{4}{c}{C-GQA} \\
\cmidrule(lr){2-5}
\multicolumn{1}{l}{Method} & AUC & HM & Seen & Unseen \\
\midrule
Composition Branch & 2.7 & 12.8 & 29.4 & 11.4 \\
%+ Direct Decoupling & 3.1 & 13.5 & 30.0 & 12.7 \\
+ Only A2O & 3.1 & 13.4 & 30.3 & 12.5 \\
+ Only O2A & 3.0 & 13.4 & \textbf{30.5} & 12.5 \\
+ A2O and O2A  & \textbf{3.4} & \textbf{14.4} & 30.4 & \textbf{13.4} \\
\bottomrule
\end{tabular}
\end{adjustbox}
\label{PPCB ab}
\end{table}

% \begin{table}[t]
% \centering
% \vspace{-0.4cm}
% \caption{Ablation study on ParamCls for C-GQA.}
% \begin{adjustbox}{width=0.95\columnwidth}
% \begin{tabular}{lcccc}
% \toprule
% \multicolumn{1}{l}{Dataset~→} & \multicolumn{4}{c}{C-GQA} \\
% \cmidrule(lr){2-5} Method~↓ & AUC & HM & Seen & Unseen \\
% \midrule
% All NonPaCls & 3.0 & 13.3 & 29.9 & 12.3 \\
% All ParamCls & 2.8 & 12.9 & 29.3 & 11.7 \\
% NonPaCls+ ParamCls & 2.7 & 12.8 & 28.9 & 11.9 \\
% ParamCls + NonPaCls (Ours) & \textbf{3.4} & \textbf{14.4} & \textbf{30.4} & \textbf{13.4} \\
% \bottomrule
% \end{tabular}
% \end{adjustbox}
% \label{CEC ab}
% \end{table}

\begin{table}[t]
\centering
\vspace{-0.3cm}
\caption{Ablation study on ParamCls in C-GQA.}
\begin{adjustbox}{width=\columnwidth}
\begin{tabular}{ccccccc}
\toprule
%\multicolumn{1}{l}{Dataset~→} & \multicolumn{4}{c}{C-GQA} \\
Model & A2O \& O2A & Composition & AUC & HM & Seen & Unseen \\
\midrule
$M_1$ & NonPaCls & NonPaCls & 3.0 & 13.3 & 29.9 & 12.3 \\
$M_2$ & ParamCls & ParamCls & 2.8 & 12.9 & 29.3 & 11.7 \\
$M_3$ & NonPaCls & ParamCls & 2.7 & 12.8 & 28.9 & 11.9 \\
$M_4$ & ParamCls & NonPaCls & \textbf{3.4} & \textbf{14.4} & \textbf{30.4} & \textbf{13.4} \\
\bottomrule
\end{tabular}
\end{adjustbox}
\label{CEC ab}
\end{table}

\begin{figure}[t]
  \centering
  \includegraphics[width=\columnwidth]{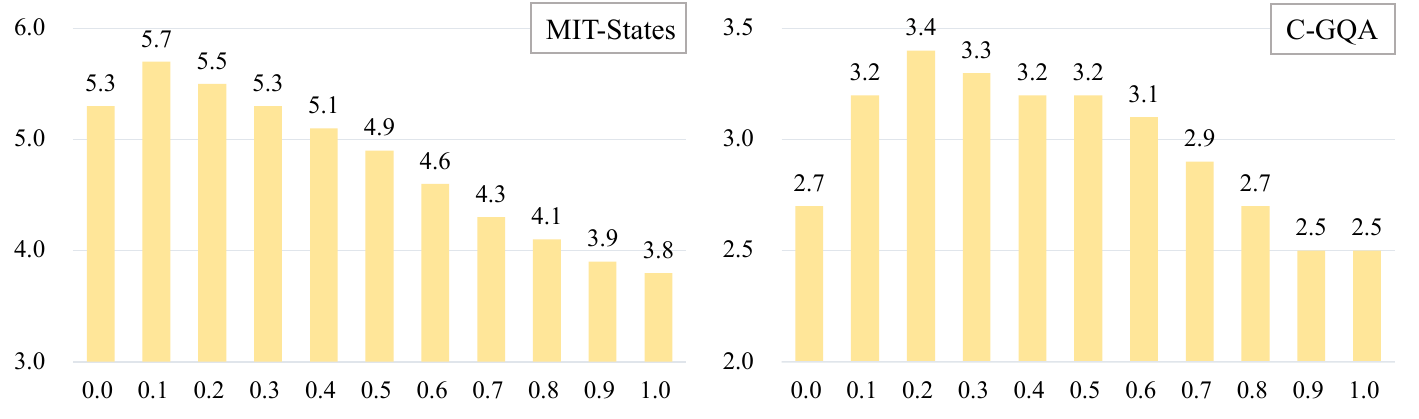}
  \vspace{-0.85cm}
   \caption{Impact of $\beta$ on AUC accuracy for two datasets. 
   % The horizontal and vertical axes represent the values of $\beta$ and AUC accuracy, respectively.
   }
   \label{fig: beta}
\end{figure}

\begin{figure}[t]
  \centering
  \includegraphics[width=\columnwidth]{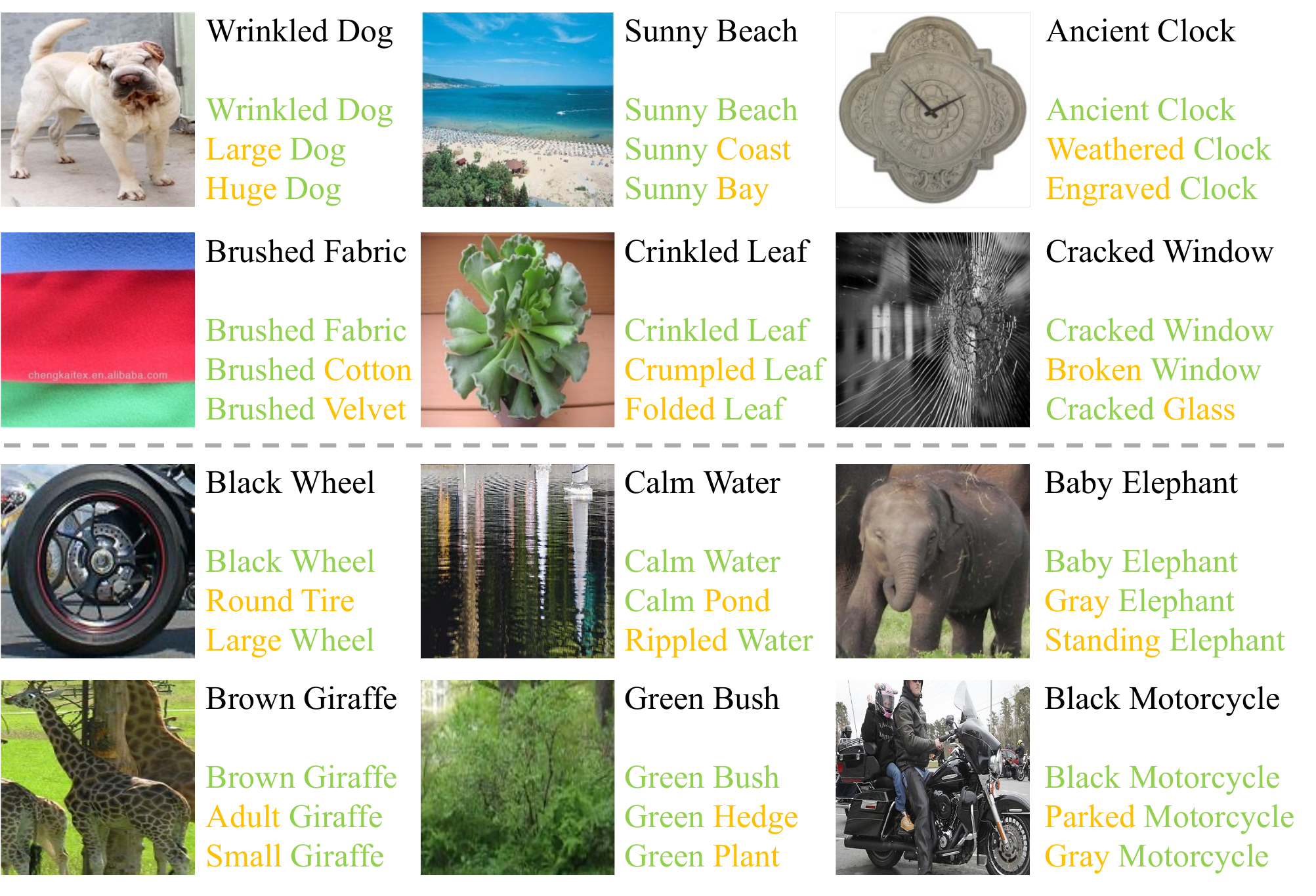}
  \vspace{-0.85cm}
   \caption{Qualitative Results on MIT-States (Top) and C-GQA (Bottom). We present Top-3 prediction candidates using CSCNet, where \textcolor[RGB]{144,209,106}{green} indicates correct predictions and \textcolor[RGB]{255,191,54}{yellow} indicates incorrect predictions.
   }
   \label{fig: qr}
\end{figure}

\subsection{Ablation Study and Analysis}
% We conduct ablation experiments on C-GQA to assess the effectiveness of A2O, O2A and ParamCls.

\textbf{Effectiveness of A2O and O2A.}
First of all, we implement a basic model with the composition branch only. Then we add more branches on top of this basic model to study the effectiveness of A2O and O2A. 
As shown in Tab.~\ref{PPCB ab}, 
equipping the model with either A2O or O2A improves all the metrics, compared to using the composition branch only.
Moreover, when both A2O and O2A are deployed, the model outperforms the counterparts with only A2O or O2A and achieves the best performance.
It verifies that the exploration of A-O dependency benefits disentangled representation learning.

% Next, we analyze the effectiveness of the CMC and present the results in Tab.~\ref{CEC ab}.
% The optimal performance is achieved when only primitives branches utilize the CMC, indicating the validity of the addition of CMC compared with the first row (applying cosine classifier in all branches). However, when cosine classifier is applied to the primitives branches and a CMC is used on the compositions branch (the third row), the model's performance falls behind the results achieved when using either cosine classifier or CMC exclusively. This reveals that CMC is not suitable for the compositions branch, as we discussed in Section 2.4.

\textbf{Effectiveness of parametric classifier.}
As discussed in Section 2.3, ParamCls is suitable for both A2O and O2A branches, but not for the composition branch. 
To prove it, we conduct this ablation experiment as shown in Tab.~\ref{CEC ab}.
When comparing $M_4$ with $M_1$, we can observe that using ParamCls for A2O and O2A, instead of non-parametric classifier (\ie NonPaCls), brings considerable boosts for all the metrics.
However, when ParamCls is applied to the composition branch,
\ie $M_2$ and $M_3$, the performance is even inferior to that of $M_1$.
These results further verify our explanation about why ParamCls is not suitable for the composition branch.

\textbf{Impact of $\beta$ on inference.}
Inspired by \cite{punan1,punan2}, this experiment studies the impact of tuning the hyper-parameter $\beta$ in Eq.~\ref{equ: score} during inference. 
As shown in Fig.~\ref{fig: beta}, the AUC accuracy exhibits a trend of initially increasing and then decreasing on both datasets. 
Finally, CSCNet achieves the best performance when $\beta$ is 0.1 for MIT-States and 0.2 for C-GQA, respectively.

\subsection{Qualitative Analysis}
Fig.~\ref{fig: qr} shows several Top-3 prediction candidates of CSCNet on MIT-States (Top) and C-GQA (Bottom). We can see that the first candidate precisely match the ground-truth. In addition, the second and third candidates still contain the correct primitives (attributes or objects) or their synonyms. It is evident that the results can still accurately describe the image. 
% Handling the challenge of multiple labels is an unsolved yet promising problem for CZSL.

\section{Conclusion}
In this work, in order to model the contextual dependency, 
we have proposed a novel A-O disentanglement framework named Class-specified Cascaded Network (CSCNet). Specifying the result of the attribute classification as a priori, we further guide the object classification
in a cascaded fashion.
Simultaneously, we classify attributes under the guidance of object classification. In addition, we employ a parametric classifier to learn optimal matching scores. Experiments and ablation studies on two datasets demonstrate the superiority of our method.
In future work, it is promising to apply CSCNet to a open-world CZSL setting.

\vfill\pagebreak

\bibliographystyle{IEEEbib}
\bibliography{strings,refs}

\begin{thebibliography}{10}

\bibitem{RedWine}
Ishan Misra, Abhinav Gupta, and Martial Hebert,
\newblock ``From red wine to red tomato: Composition with context,''
\newblock in {\em {Proc. of the CVPR}}, 2017, pp. 1160--1169.

\bibitem{AFGC}
Kun Wei, Muli Yang, Hao Wang, Cheng Deng, and Xianglong Liu,
\newblock ``Adversarial fine-grained composition learning for unseen
  attribute-object recognition,''
\newblock in {\em {Proc. of the ICCV}}, 2019, pp. 3740--3748.

\bibitem{causal}
Yuval Atzmon, Felix Kreuk, Uri Shalit, and Gal Chechik,
\newblock ``A causal view of compositional zero-shot recognition,''
\newblock in {\em {the NeurIPS}}, 2020.

\bibitem{Dual-Stream}
Yanhua Yang, Rui Pan, Xiangyu Li, Xu~Yang, and Cheng Deng,
\newblock ``Dual-stream contrastive learning for compositional zero-shot
  recognition,''
\newblock {\em IEEE Transactions on Multimedia}, 2023.

\bibitem{punan4}
Nan Pu, Wei Chen, Yu~Liu, Erwin~M. Bakker, and Michael~S. Lew,
\newblock ``Dual gaussian-based variational subspace disentanglement for
  visible-infrared person re-identification,''
\newblock in {\em {ACM MM}}, 2020, pp. 2149--2158.

\bibitem{IVR}
Tian Zhang, Kongming Liang, Ruoyi Du, Xian Sun, Zhanyu Ma, and Jun Guo,
\newblock ``Learning invariant visual representations for compositional
  zero-shot learning,''
\newblock in {\em {Proc. of the ECCV}}, 2022, pp. 339--355.

\bibitem{OADis}
Nirat Saini, Khoi Pham, and Abhinav Shrivastava,
\newblock ``Disentangling visual embeddings for attributes and objects,''
\newblock in {\em {Proc. of the CVPR}}, 2022, pp. 13658--13667.

\bibitem{DECA}
Muli Yang, Chenghao Xu, Aming Wu, and Cheng Deng,
\newblock ``A decomposable causal view of compositional zero-shot learning,''
\newblock {\em IEEE Transactions on Multimedia}, pp. 1--11, 2022.

\bibitem{BMP-Net}
Ziwei Xu, Guangzhi Wang, Yongkang Wong, and Mohan~S. Kankanhalli,
\newblock ``Relation-aware compositional zero-shot learning for
  attribute-object pair recognition,''
\newblock {\em IEEE Transactions on Multimedia}, vol. 24, pp. 3652--3664, 2022.

\bibitem{punan3}
Nan Pu, Wei Chen, Yu~Liu, Erwin~M. Bakker, and Michael~S. Lew,
\newblock ``Lifelong person re-identification via adaptive knowledge
  accumulation,''
\newblock in {\em {Proc. of the CVPR}}, 2021, pp. 7901--7910.

\bibitem{SeCoNet}
Aditya Panda, Bikash Santra, and Dipti~Prasad Mukherjee,
\newblock ``Isolating features of object and its state for compositional
  zero-shot learning,''
\newblock {\em IEEE Transactions on Emerging Topics in Computational
  Intelligence}, pp. 1--13, 2023.

\bibitem{CANet}
Qingsheng Wang, Lingqiao Liu, Chenchen Jing, Hao Chen, Guoqiang Liang, Peng
  Wang, and Chunhua Shen,
\newblock ``Learning conditional attributes for compositional zero-shot
  learning,''
\newblock in {\em {Proc. of the CVPR}}, 2023, pp. 11197--11206.

\bibitem{w2v}
Tom{\'{a}}s Mikolov, Kai Chen, Greg Corrado, and Jeffrey Dean,
\newblock ``Efficient estimation of word representations in vector space,''
\newblock in {\em {Proc. of the ICLR}}, 2013.

\bibitem{AttrAsOp}
Tushar Nagarajan and Kristen Grauman,
\newblock ``Attributes as operators: Factorizing unseen attribute-object
  compositions,''
\newblock in {\em {Proc. of the ECCV}}, 2018, pp. 172--190.

\bibitem{TaskDriven}
Senthil Purushwalkam, Maximilian Nickel, Abhinav Gupta, and Marc'Aurelio
  Ranzato,
\newblock ``Task-driven modular networks for zero-shot compositional
  learning,''
\newblock in {\em {Proc. of the ICCV}}, 2019, pp. 3592--3601.

\bibitem{Symnet}
Yong{-}Lu Li, Yue Xu, Xiaohan Mao, and Cewu Lu,
\newblock ``Symmetry and group in attribute-object compositions,''
\newblock in {\em {Proc. of the CVPR}}, 2020, pp. 11313--11322.

\bibitem{Compcos}
Massimiliano Mancini, Muhammad~Ferjad Naeem, Yongqin Xian, and Zeynep Akata,
\newblock ``Open world compositional zero-shot learning,''
\newblock in {\em {Proc. of the CVPR}}, 2021, pp. 5222--5230.

\bibitem{CGE}
Muhammad~Ferjad Naeem, Yongqin Xian, Federico Tombari, and Zeynep Akata,
\newblock ``Learning graph embeddings for compositional zero-shot learning,''
\newblock in {\em {Proc. of the CVPR}}, 2021, pp. 953--962.

\bibitem{SRA}
T.~Guo, J.~Liang, and G.~Xie,
\newblock ``Swap-reconstruction autoencoder for compositional zero-shot
  learning,''
\newblock in {\em {Proc. of the ICME}}, 2023, pp. 438--443.

\bibitem{SCEN}
Xiangyu Li, Xu~Yang, Kun Wei, Cheng Deng, and Muli Yang,
\newblock ``Siamese contrastive embedding network for compositional zero-shot
  learning,''
\newblock in {\em {Proc. of the CVPR}}, 2022, pp. 9326--9335.

\bibitem{MIT}
Phillip Isola, Joseph~J. Lim, and Edward~H. Adelson,
\newblock ``Discovering states and transformations in image collections,''
\newblock in {\em {Proc. of the CVPR}}, 2015, pp. 1383--1391.

\bibitem{ADE}
Shaozhe Hao, Kai Han, and Kwan-Yee~K. Wong,
\newblock ``Learning attention as disentangler for compositional zero-shot
  learning,''
\newblock in {\em {Proc. of the CVPR}}, 2023, pp. 15315--15324.

\bibitem{resnet}
Kaiming He, Xiangyu Zhang, Shaoqing Ren, and Jian Sun,
\newblock ``Deep residual learning for image recognition,''
\newblock in {\em {Proc. of the CVPR}}, 2016, pp. 770--778.

\bibitem{ImageNet}
Jia Deng, Wei Dong, Richard Socher, Li{-}Jia Li, Kai Li, and Li~Fei{-}Fei,
\newblock ``Imagenet: {A} large-scale hierarchical image database,''
\newblock in {\em {Proc. of the CVPR}}, 2009, pp. 248--255.

\bibitem{fasttext}
Piotr Bojanowski, Edouard Grave, Armand Joulin, and Tom{\'{a}}s Mikolov,
\newblock ``Enriching word vectors with subword information,''
\newblock {\em Trans. Assoc. Comput. Linguistics}, vol. 5, pp. 135--146, 2017.

\bibitem{adam2015}
Diederik~P. Kingma and Jimmy Ba,
\newblock ``Adam: {A} method for stochastic optimization,''
\newblock in {\em {Proc. of the ICLR}}, 2015.

\bibitem{punan1}
Nan Pu, Yu~Liu, Wei Chen, Erwin~M. Bakker, and Michael~S. Lew,
\newblock ``Meta reconciliation normalization for lifelong person
  re-identification,''
\newblock in {\em {ACM MM}}, 2022, pp. 541--549.

\bibitem{punan2}
Nan Pu, Zhun Zhong, Nicu Sebe, and Michael~S. Lew,
\newblock ``A memorizing and generalizing framework for lifelong person
  re-identification,''
\newblock {\em {IEEE Transactions on Pattern Analysis and Machine
  Intelligence}}, vol. 45, no. 11, pp. 13567--13585, 2023.

\end{thebibliography}

\end{document}